\documentclass[runningheads]{llncs}
\usepackage{graphicx}
\begin{document}
\title{Extractive Text Summarization Using \\ Generalized Additive Models with Interactions for Sentence Selection}
\titlerunning{ETS Using GAMs with Interactions for Sentence Selection}
%
\author{Vinícius Camargo da Silva \and
João Paulo Papa  \and Kelton Augusto Pontara da Costa}
\authorrunning{V.C. Silva, J.P. Papa and K.A.P. Costa}
%
\institute{São Paulo State University - UNESP, Bauru, Brazil
\email{\{vinicius.camargo,joao.papa,kelton.costa\}@unesp.br }}
\maketitle              
\begin{abstract}

Automatic Text Summarization (ATS) is becoming relevant with the growth of textual data; however, with the popularization of public large-scale datasets, some recent machine learning approaches have focused on dense models and architectures that, despite producing notable results, usually turn out in models difficult to interpret. Given the challenge behind interpretable learning-based text summarization and the importance it may have for evolving the current state of the ATS field, this work studies the application of two modern Generalized Additive Models with interactions, namely Explainable Boosting Machine and GAMI-Net, to the extractive summarization problem based on linguistic features and binary classification.
\end{abstract}

\section{Introduction}

Nowadays, computer systems have assumed important role in providing useful information inside the increasingly amount of data generated daily. In this context, Natural Language Processing (NLP) have gained space and applicability as a result of considerable amounts of text distributed across news portals, social media and various sources. With the growth of textual data, finding exact information may be a difficult task \cite{el2020automatic}. As such, Automatic Text Summarization (ATS) is becoming relevant \cite{el2020automatic}, as it fosters automatic  strategies for building textual comprehensible summaries, ideally capable of preserving the original content and meaning \cite{moratanch2017survey} while distilling the most important information considering the user involved \cite{maybury1995generating}.

Some of the previous ATS approaches were simpler and had certain transparency, for example, using linear equations  or fuzzy rules with statistical or linguistic features to map the importance of document sentences to produce extractive summaries \cite{mutlu2019multi,afsharizadeh2018query}. In general, the simpler and the clearer the predictors used, the more directly it is possible to observe how the summary sentences are being chosen in case of extractive summarizers.

However, not all available approaches are transparent when considering 
model's decisions, especially in the machine learning context. With the popularization of public large-scale datasets, various deep learning approaches have been explored during the last years, competing for the state-of-the-art on such data. Despite their notable ability to elaborate summaries, as a problem inherited from dense architectures and language embeddings opacities \cite{danilevsky2020survey}, in practice, such models usually yield shallow or obscure interpretations about their true behaviour and decisions.

Machine learning applications have been successful in different areas based on their statistical accuracy, but often lack clarity when explaining how decisions are actually being made. The so-called XAI (or Explainable Artificial Intelligence) field focuses on the study and elaboration of more explainable AI methodologies \cite{dovsilovic2018explainable,samek2019towards,arrieta2020explainable}. Depending on the application, explainability could impact in different aspects of a model, as its lack of interpretability can undermine its trustworthiness towards users or even hide potential improvements \cite{danilevsky2020survey,arrieta2020explainable}.

Within the context of ATS, interpretable modeling relates to giving transparency to the model's summarization process, which could contribute to a better sense of the limitations and capabilities of the model, help the investigation of why the model makes mistakes or even assist in gaining insights about the problem itself. Such information could be useful, for example, for evolving approaches and for clarifying what the model actually satisfy in contrast to user's expectations.

With the interest in moving towards interpretable ATS learning models, this work aims to study the application of Generalized Additive Models with Interactions (GAMI) to the extractive summarization problem. More specifically, this work investigates training Explainable Boosting Machine (EBM) and GAMI-Net models in a binary classification fashion to later inferring the relevancy of sentences in the documents of interest.

To the best of our knowledge, this is the first application of EBMs or GAMI-Nets as the decision algorithm involved on extractive summarization. EBM and GAMI-Net models are built, respectively, on ensemble of trees and neural networks, whose main attempt is to balance intelligibility and accuracy in supervised problems combining main effects and pairwise interactions additively. The idea is to benefit from the additive formulation to make the behaviour and contributions explicit considering explanatory features and outputs, allowing intelligibility along the process. This work evaluates these models for news summarization (CNN/Dailymail) and long document summarization of scientific papers (PubMed), comparing to other machine learning algorithms and recent approaches.

\section{Background}

Automatic Text Summarization is an NLP task that grows in importance with the expansion of data in textual form and the interest in exploring it efficiently, since assisting users understanding over documents could save time and effort \cite{el2020automatic}. 

As pointed by \cite{luhn1958automatic}, the summarization process may require familiarity with the subject, which could culminate in qualified human resources dedicated to facilitating access to information. Therefore, the importance of ATS relies on the potential of reducing human efforts while accelerating reading time over text sources \cite{moratanch2017survey} provided by automatic summarization.

A common way to distinguish text summarization approaches is between extractive and abstractive modelling. On extractive summarization, the model selects parts of the source text to compose the projected summary \cite{nenkova2011automatic}, whereas on abstractive summarization the model may reuse parts of the source, but new terms and sentences are expected to appear \cite{nenkova2011automatic}.

Traditionally, extractive techniques have the advantage of not suffering from grammatical or semmantical issues in the summary \cite{nallapati2017summarunner}, as usually the approaches are based on selecting entire sentences from the source text, leading to faster and simpler methods than abstractive techniques \cite{el2020automatic}. One way to give general description of the extractive summarization process can be done through three important steps \cite{nenkova2012survey}: 
\begin{enumerate}
    \item Creating an intermediate representation of the input;
    \item Scoring sentences based on this representation; and
    \item Selecting sentences to compose the summary.
\end{enumerate}
Supervised machine learning algorithms emerged as an option
for modelling extractive summarization \cite{el2020automatic}, offering statistical performance at the price of requiring a dataset of considerable size
\cite{wong2008extractive}.

Considering the interest on fully explainable models \cite{arrieta2020explainable,danilevsky2020survey,samek2020toward}, bringing explainability to ATS is becoming a necessity \cite{sarkhel2020interpretable}.
Being able to produce summaries through interpretable approaches may be an important step for machine learning-based models; however, literature focused on this topic is still scarce. 

Recently, the authors in \cite{ghodratnama2020extractive} combined supervised and unsupervised learning into a model called ExDoS that learns features weights as part of an extractive summarization approach. Such weights are discussed by the authors as a form to indicate the learned importances of the features, which may help the interpretation of how the model is deciding to select important sentences and avoid unimportant ones when elaborating summaries.

With a similar intention, in this work, we present an attempt to build interpretable extractive summarization models by taking advantage of Generalized Additive Models with Interactions, relying on their transparency concerning features and outputs, to give intelligibility to the sentence selection process. 

\subsection{Generalized Additive Models with Pairwise Interactions}

Generalized Additive Models (GAMs) \cite{hastie1987generalized} are statistical approaches that have gained interest for their ability to be intelligible, appealing to their additive formulation and human intuition to provide interpretability in supervised problems. 

These models have the form of Equation \ref{gam-equation}, working as middle ground between linear models (e.g., linear regression) and full-complexity models (e.g., ensembles of trees) \cite{lou2012intelligible}, where the motivation could be, for example, obtaining predictors more accurate than the former while staying more intelligible than the latter, fitting individual features with non-linear functions that are combined additively \cite{lou2012intelligible}:

\begin{equation}
    g(y) = \sum_{i \in N} f_i(x_i) \label{gam-equation},
\end{equation}
where $x_i$ is the $i$-th feature given a set $N$ of features, $f$ is called \textit{shape function}, and $g$ is called \textit{link function}.

On additive models, the contribution of each feature map towards final decisions can be seen more clearly than in dense ones. This property allows GAM's learned shape functions to be visualized  and their outputs to be investigated for individual or group of predictions, providing interpretability over the model and its decisions.  Moreover, as GAMs can assume non-linear behaviour through their shape functions, they may fit a wider variety of problems when compared to linear models.

Recently, some works were focused on effectively building more powerful GAMs with the addition of modern machine learning strategies and supplementing the original univariate shape functions with a usual restricted set $K$ of pairwise interactions (bivariate shape functions), increasing the accuracy of the final model while maintaining some intelligibility, resulting in the following models:

\begin{equation}
    g(y) = \sum_{i \in N} f_i(x_i) + \sum_{(i,j) \in K} f'_{ij}(x_i, x_j), \label{gam-pair-equation}
\end{equation}
where $f'$ is a pairwise interaction. 

In this work, two representatives of those algorithms are applied to the extractive summarization problem, i.e., EBM and GAMI-Net.

\subsubsection{Explainable Boosting Machine}

Explainable Boosting Machines (also known as Generalized Additive Model plus Interactions (GA$^2$M) \cite{lou2013accurate}) are modern tree-based GAMs with pairwise interactions based on ensembles, which achieved accuracy comparable to full-complexity models while keeping interpretability similar to former GAMs \cite{lou2012intelligible,lou2013accurate,nori2019interpretml}. 

Given a supervised problem, tree-based GAMs can be learned upon residuals using gradient boosting, cycling through the features and improving shape functions iteratively \cite{lou2012intelligible}. Using this algorithm with bagged trees have led to even better results on different regression and binary classification datasets \cite{lou2012intelligible}. 

Another inprovement was addresed by \cite{lou2013accurate}, encompassing the inclusion of pairwise interactions into tree-based GAMs. Considering computational cost, the authors propose an approach based on firstly building an additive model with only univariate shape functions, and then ranking and selecting a fixed number of pairwise interactions that are fit on the residuals. The model that wraps those improvements was later called \textit{Explainable Boosting Machine} by \cite{nori2019interpretml}.

\subsubsection{GAMI-Net}

GAMI-Net is an interpretable neural network based on
GAMs with structured interactions \cite{yang2021gami}. The architecture consists in additive subnetworks with multiple hidden layers, each of which capturing a different shape function of Equation \ref{gam-pair-equation}. The idea is to produce a model that keeps interpretability aspects of GAMs while relying on the power of deep neural networks to model non-linear behavior of shape functions.

The authors in \cite{yang2021gami} proposed an adaptive training algorithm using mini-batch gradient descent that fits main effects and pairwise interactions in separated stages. Firstly, the algorithm train the main subnetworks, pruning the trivial ones according to their contributions. Secondly, the algorithm selects a fixed number of pairwise interactions using the ranking procedure proposed by \cite{lou2013accurate} and fit their respective subnetworks on the residuals, pruning trivial ones as in the first stage. Lastly, in a third and final stage, all the network parameters are fine-tuned together. 

Moreover, GAMI-Net is designed to preserve sparsity, heredity and marginal clarity considering the main effects and pairwise interactions by keeping only non-trivial shape functions, including pairwise interactions only if at least one of their parent main effects are kept and putting a regularization factor that enforces main effects and pairwise interactions to be more identifiable, which is intended to contribute to the model's interpretability \cite{yang2021gami}.

\section{Proposed approach}
\label{approach}

Let $D = \{s_0, ..., s_n\}$ be a document composed of a sequence of sentences $s_i$. The goal of our extractive summarization process is to obtain a sequence $S$ of the most relevant sentences in $D$, where $S$ is limited in size to be shorter than $D$. 

We train EBM and GAMI-Net models using a set of six features from the literature to be able to rank and select sentences to compose $S$. The approach consists in training a model to decide which sentences are summary-worthy and then using such model to select the most important sentences in the input document to compose summaries. The simplicity of the explored features should contribute to both model efficiency and intelligibility. The approach is divided in preprocessing, feature extraction and sentence scoring and selection, as follows.

\subsection{Preprocessing}
\label{preprocessing}

The preprocessing step is responsible for turning raw documents into sequences of sentences, capturing useful information for future feature extraction. The process starts by segmenting raw documents into text sentences, which are split in tokens. Then, the process follows by removing punctuation and stopwords, word tagging and stemming. After preprocessing, sentences correspond to lists of their respective words that are forwarded to the feature extraction step.

We perform most of the NLP preprocessing with the Python library spaCy \cite{montani2021spacy}, with the exception of stemming step which is done using the NLTK SnowballStemmer \cite{bird2009natural}.

\subsection{Feature extraction}
\label{feature-extraction}

After sentences are preprocessed, six different features are extracted from sentences to become inputs vectors $x = \{x_1, x_2, ..., x_6\}$ that are used to train and predict. The feature computations are formulated as described below:

\subsubsection{TF-ISF}

TF-ISF is a variant of the TF-IDF method applied at sentence level for text summarization \cite{oliveira2016assessing,mutlu2019multi}. The idea is to compute a score to each sentence based on term importance and descriptiveness inside the document \cite{oliveira2016assessing}, which are measured by term frequency (TF) and inverse sentence frequency (ISF) of the terms. We use bigrams TF-ISF, so each sentence $s_i$ of a document receives a salience score (Equation \ref{tfisf}) based on its term bigrams $b$:

\begin{equation}
w(s_i)  = \sum_{j = 1}^{J_i} \left[ \textrm{F}(b_{j}) \times \textrm{log} \left( \frac{n}{n_{b_j}} \right) \right],
\end{equation}

\begin{equation}
x_1 (s_i) = \frac{w(s_i)}{\max  (w(s_i))}\label{tfisf}.
\end{equation}
where $\textrm{F}(b)$ is the frequency of $b$ in the document,  $n$ is the number of sentences in the document, $n_{b}$ is the number of sentences of the document in which $b$ occurs and $J_i$ is the number of bigrams $s_i$.
\subsubsection{Position}
Depending on a document type, how early or how late sentences appear may give important information about their relevancy \cite{ferreira2013assessing,oliveira2016assessing}.
The position feature (Equation \ref{position}) represents the sentence position inside the document, where $p_i$ is the position of sentence $s_i$:
\begin{equation}
x_2(s_i) = \frac{p_i}{n} \label{position}.
\end{equation}

\subsubsection{Length}
The length feature (Equation \ref{length}) is calculated based on the length of sentence $s_i$ in terms of the maximum sentence length. The length feature allows the model to learn the relationship between sentence length and relevancy \cite{oliveira2016assessing,mutlu2019multi}:

\begin{equation}
x_3(s_i) = \frac{\textrm{number of terms in sentence} \ s_i}{\max (\textrm{number of terms in a sentence})} \label{length}.
\end{equation}

\subsubsection{Proper Nouns and Numerical} 
The individual ratio of proper noun and numerical terms in the sentence $s_i$ may indicate the presence of relevant information \cite{oliveira2016assessing}. We calculate these features as follows:

\begin{equation}
x_4(s_i) = \frac{\textrm{number of proper nouns in }s_i}{\textrm{number of terms in }s_i}  \label{pn} \ \ \ \textrm{and}
\end{equation}

\begin{equation}
x_5(s_i) = \frac{\textrm{number of numerical terms in }s_i}{\textrm{number of terms in }s_i}  \label{num}.
\end{equation}

\subsubsection{Sentence-sentence similarity}
The sentence-sentence similarity denotes how close a sentence is to other sentences in the  document \cite{mutlu2019multi}. We calulate this feature using cosine similarity $c$ as in Equation \ref{sim}: 

\begin{equation}
x_6(s_i) = \frac{\sum_{j=1}^n c(s_i, s_j)}{\max (\sum_{j=1}^n c(s_k, s_j))}, \ \ \ i \neq j. \label{sim}
\end{equation}

\subsection{Sentence scoring and selection}
\label{sentence-scoring}

In order to select sentences to produce summaries, as various works did in the past, we interpret the Extractive Summarization problem as a binary classification one, where the model is trained to decide between exclusion and inclusion of individual sentences in the summary.

The procedure consists in minimizing the binary loss in a supervised learning approach where the sentences' feature vectors and the labels that classify whether they are present or not in their respective document summaries are used to train the model. 

After training, we use the model's ability to distinguish between ``important" and ``unimportant'' sentences to, given an input document, score and select appropriate sentences among the others. The scoring process consists in obtaining the probability of a sentence being part of the summary given its feature vector $x$, for each of the sentences in the document. Then, similarly to \cite{nallapati2017summarunner,kedzie2018content,xiao2019extractive}, the document summary is obtained by ranking and selecting top sentences using that probability, respecting the length compression  limit.

\section{Experiments}

\subsection{Datasets}
In this work, we compare EBM and GAMI-Net models with other approaches on two public text summarization datasets, namely CNN/Dailymail \cite{hermann2015teaching,see2017get} and Pubmed \cite{cohan2018discourse}. The CNN/Dailymail summarization dataset consists of pairs of news articles and their main highlights constituting 312K documents. This dataset have been widely adopted on recent ATS works, especially by Recurrent Neural Network and Transformer-based approaches. We use the non-anonymized version of this dataset \cite{see2017get}.

The PubMed dataset is a collection of scientific papers totaling 133K documents in which the abstract section is used as the summary references. This dataset have been used for evaluating long document summarization approaches as both document and summaries are usually longer than in news datasets \cite{xiao2019extractive}. 

As mentioned in Section \ref{approach}, our approach uses individual sentences as the input instances for training. Considering that initially both datasets only possess abstractive summaries, we needed extractive oracle labels to train our models. Recently, different authors obtained those labels using automatic heuristics while working with these datasets \cite{nallapati2017summarunner,kedzie2018content,liu2019fine,xiao2019extractive}, which we adopt here. For CNN/Dailymail, we generated labels using the scripts provided by \cite{liu2019fine}\footnote{https://github.com/nlpyang/BertSum}, and, for Pubmed, we utilized labels extracted and made public by \cite{xiao2019extractive}\footnote{https://github.com/Wendy-Xiao/Extsumm\_local\_global\_context}. Moreover, we adopted random undersampling to handle label inbalance during the training step.

\subsection{Model Comparison}

We compare EBM and GAMI-Net with some recent baselines, most of which produced by deep neural architectures, in the sense of outlining the summarization ability in contrast to these models, despite the notable differences in terms of interpretability. Tables \ref{table-cnndm} \ref{table-pubmed} present the results and reporting authors in CNN/Dailymail and Pubmed datasets, respectively.

Additionally, we compare EBM and GAMI-Net models with other supervised machine learning classifiers, i.e., Logistic Regression (LR), Random Forest (RF) and XGBoost, using the exact same fashion and features described in Section \ref{approach}. Each technique was trained and tested ten times and the average scores are considered for comparison purposes. 

Overall approaches are evaluated using the ROUGE score metric \cite{lin2004rouge} considering its broad adoption for extractive summarization systems. Also, we evaluate the sentence selection ability of the supervised classifiers computing summary F1 scores based on the oracle labels. Moreover, Lead baseline correspond to selecting the first sentences present in the documents as the summaries (respecting each dataset summary-length limit) and Oracle denote the pre-obtained oracle labels' scores. Our ROUGE scores were obtained using pyrouge\footnote{https://pypi.org/project/pyrouge/}, a python wrapper of the original ROUGE-1.5.5 scripts and, concerning summary sizes, CNN/Dailymail summaries were limited to three sentences \cite{zhong2020extractive} while Pubmed summaries were limited to 200 words \cite{xiao2019extractive}.

\begin{table}[!htbp]
\caption{Results on CNN/Dailymail dataset.}
\label{table-cnndm}
\begin{center}
\begin{tabular}{|c|c|c|c|c|c|c|}
\hline
&&&\multicolumn{3}{|c|}{\textbf{ROUGE F (\%)}}& \textbf{F1}\\
\cline{4-6} 
\textbf{Models} & \textbf{Type} & \textbf{Interpretability} & \textbf{\textit{R-1}}& \textbf{\textit{R-2}}& \textbf{\textit{R-L}}& \textbf{(\%)} \\
\hline
\hline
Lead & -- & -- & 40.12 & 17.54 & 36.30 & -- \\
Oracle & -- & -- & 56.09 & 33.67 & 52.21 & -- \\
\hline
\hline
P-Gen \cite{see2017get} & Abs. & Low & 39.53 & 17.28 & 36.38 & -- \\
BART + RD \cite{wu2021r} & Abs. & Low & 44.51 & 21.58 & 41.24 & -- \\
SummaRuNNer \cite{ghodratnama2020extractive} & Ext. & Low & 39.90 & 16.30 & 35.10 & -- \\
MatchSum \cite{zhong2020extractive} & Ext. & Low & 44.41 & 20.86 & 40.55 & -- \\
ExDoS \cite{ghodratnama2020extractive} & Ext. & High & 42.1 & 18.9 & 37.7 & -- \\
\hline
\hline
LR & Ext. & High & 38.51 & 16.66 & 34.74 & 31.96 \\
RF & Ext. & Low & 39.46 & 17.34 & 35.71 & 32.71 \\
XGBoost & Ext. & Low & 39.58 & 17.50 & 35.82 & 33.52 \\
\hline
\hline
EBM & Ext. & High & 39.48 & 17.40 &	35.74 & 33.40 \\
GAMI-Net & Ext. & High & 39.52 & 17.42 & 35.76 & 33.40 \\
\hline
\end{tabular}
\end{center}
\end{table}

\subsection{Discussion \label{results}}

\sloppy
As denoted in Tables \ref{table-cnndm} and \ref{table-pubmed}, EBM and GAMI-Net achieved similar results on both datasets. Considering ROUGE scores, GAMI-Net is ahead on CNN/Dailymail while EBM is superior on Pubmed, for less than 0.1 for each variant.  

\begin{table}[htbp]
\caption{Results on PubMed dataset.}
\label{table-pubmed}
\begin{center}
\begin{tabular}{|c|c|c|c|c|c|c|}
\hline
&&&\multicolumn{3}{|c|}{\textbf{ROUGE F (\%)}}& \textbf{F1} \\
\cline{4-6} 
\textbf{Models} & \textbf{Typt} & \textbf{Interpretability} & \textbf{\textit{R-1}}& \textbf{\textit{R-2}}& \textbf{\textit{R-L}}&\textbf{(\%)} \\
\hline
\hline
Lead & -- & -- & 37.38 & 12.65 & 33.71 & -- \\
Oracle & -- & -- & 55.37 & 26.31 & 49.07 & -- \\
\hline
\hline
Discourse-aware \cite{cohan2018discourse}& Abs. & Low & 38.93 & 15.37 & 35.21 & -- \\
SummaRuNNer \cite{xiao2019extractive} & Ext. & Low & 43.89 & 18.78 & 30.36 & -- \\
ExtSum-LG \cite{xiao2020systematically} & Ext. & Low & 45.18 & 20.20 & 40.72 & -- \\
ExtSum-LG+MMR-S+ \cite{xiao2020systematically} & Ext. & Low & 45.39 & 20.37 & 40.99 & -- \\
\hline
\hline
LR & Ext. & High & 38.07 & 12.70 & 32.94 & 27.61 \\
RF & Ext. & Low & 40.16 & 14.13 & 34.98 & 30.26 \\
XGBoost & Ext. & Low & 40.16 & 14.18 & 34.97 & 30.86\\
\hline
\hline
EBM & Ext. & High & 39.86 & 13.96 & 34.65 & 30.70 \\
GAMI-Net & Ext. & High & 39.78  & 13.92 & 34.57 & 30.76 \\
\hline
\end{tabular}
\end{center}
\end{table}

As shown in Table \ref{table-cnndm}, comparing to other approaches,  EBM and GAMI-Net models were able to compete with SummaRunner \cite{nallapati2017summarunner} and Pointer-Generator \cite{see2017get} networks -- two of the earliest deep summarization architectures proposed in the past, surpassing the former concerning R-L and both of them considering R-2 on the CNN/Dailymail dataset. On the other hand, they could not overcome BART+RD \cite{wu2021r} and MatchSum \cite{zhong2020extractive} (Transformer-based models) or ExDoS \cite{ghodratnama2020extractive} in terms of scores. On the Pubmed dataset, as Table \ref{table-pubmed} shows, EBM and GAMI-Net achieved higher R-L scores than SummaRuNNer, but failed to compete with ExtSum-LG \cite{xiao2019extractive} and ExtSum-LG+MMR-S+ \cite{xiao2020systematically}.

\begin{figure}[!bp]
\centerline{\includegraphics[width=0.85\textwidth]{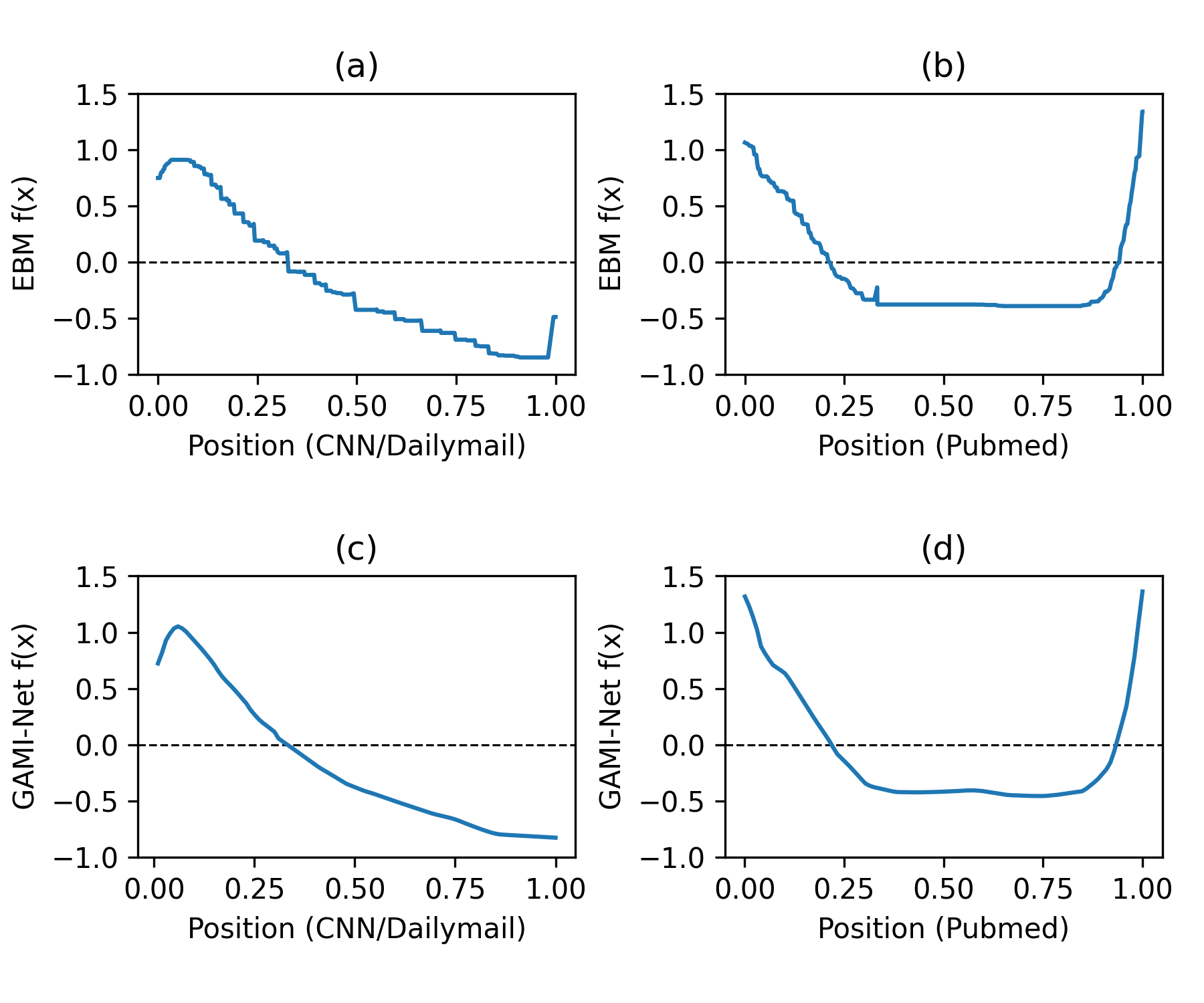}}
\caption{\textit{Position} shape function.}
\label{position_fig}
\end{figure}

Although EBM and GAMI-Net fail to approximate the ROUGE scores of the most advanced summarization approaches, they are able to provide higher levels of transparency on how predictions are being built. For Extractive Summarization, this could be useful for clarifying what is being considered by the models while deciding the importance of the sentences. Figure \ref{position_fig} presents the plot of  shape functions built upon the \textit{Position} feature (Equation \ref{position}) given the respective model and dataset, where the horizontal axes represent feature values and the vertical axes denotes the corresponding shape function outputs. In practice, this kind of view allows further investigation of the summarization models, for example, inspecting how training models on different document types may be affecting the learning.

As Figure \ref{position_fig} shows, the shape functions fit on CNN/Dailymail (Figures \ref{position_fig}a and \ref{position_fig}c) tend to give higher outputs to sentences in the beginning of the documents, while on Pubmed (Figures \ref{position_fig}b and \ref{position_fig}d) sentences both at the beginning and the end of the documents are likely to be prioritized by the models. In this case, a possible assumption is that models trained on CNN/Dailymail (news type) are likely to avoid sentences at the end of the documents when looking for relevant sentences, which is not the case for all types of documents, as seems to happen for models trained on Pubmed (scientific article type/long document summarization). EBM and GAMI-Net ability to capture such properties intrinsically and grant the possibility of further exploration can be seen as a great benefit when comparing to full-complexity models.

\begin{figure}[!b]
\centerline{\includegraphics[width=0.9\textwidth]{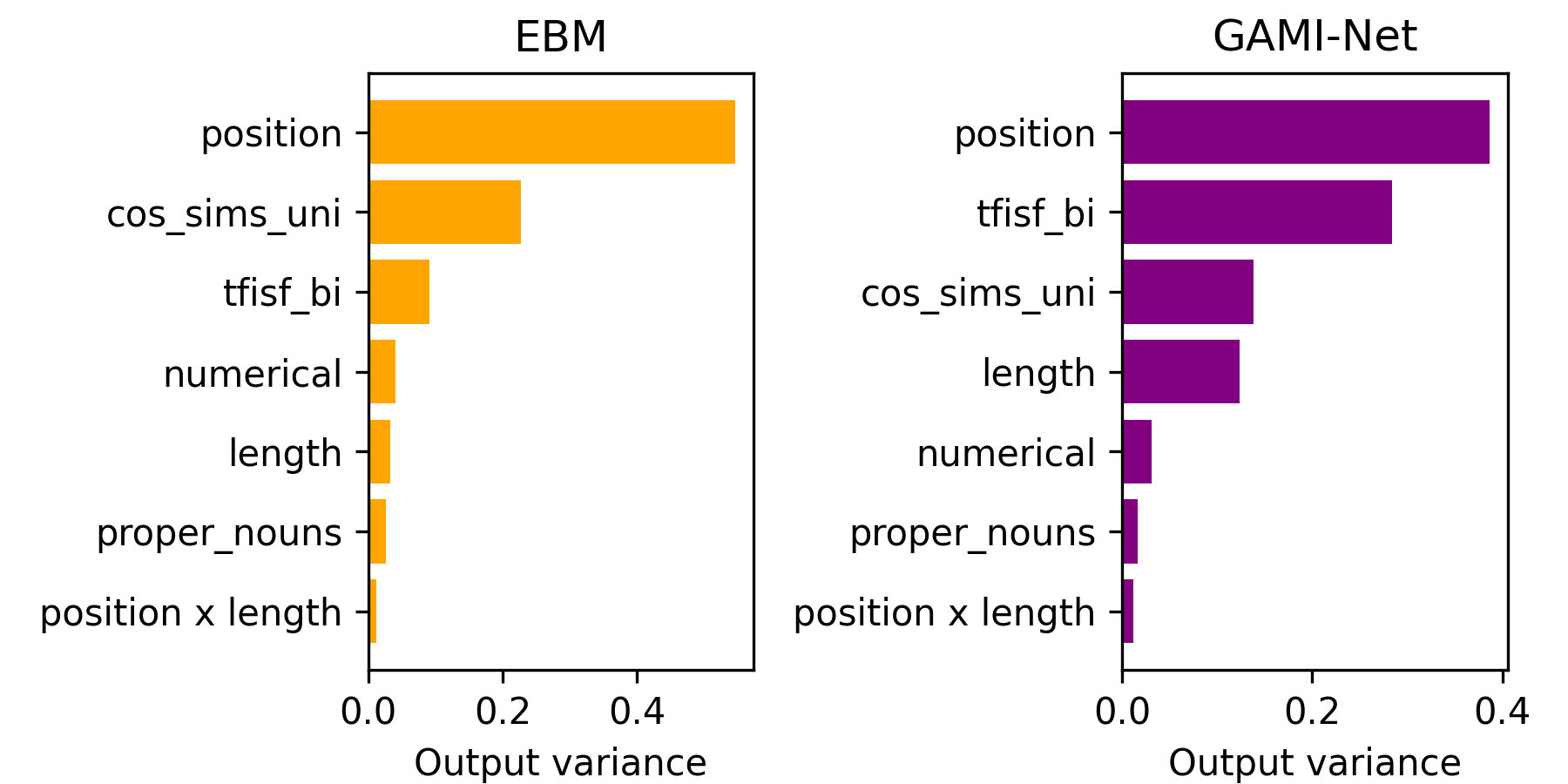}}
\caption{Top-7 importance ratios on CNN/Dailymail dataset.}
\label{overall_cnndm}
\end{figure}

\begin{figure}[!t]
\centerline{\includegraphics[width=0.9\textwidth]{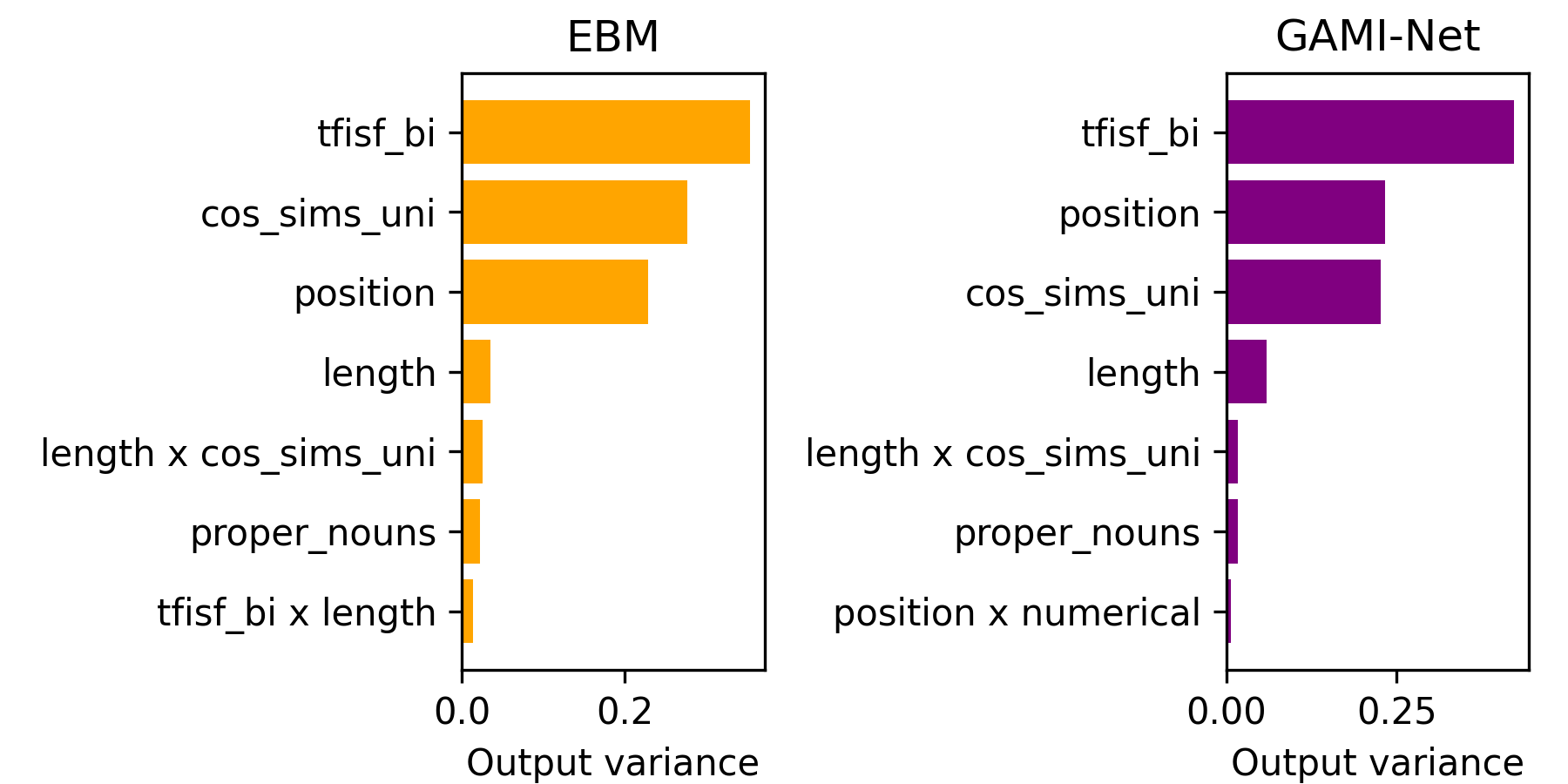}}
\caption{Top-7 importance ratios on Pubmed dataset.}
\label{overall_pubmed}
\end{figure}

Additionally, EBM and GAMI-Net feature outputs can be easily investigated for single predictions or a group of samples, helping better understanding of feature contributions while producing the summaries. Figures \ref{overall_cnndm} and \ref{overall_pubmed} show the most contributive feature effects quantified by their variation \cite{yang2021gami} on test set considering CNN/Dailymail and Pubmed datasets, respectively, where \textit{Sentence-Sentence Similarity}, \textit{TF-ISF} and \textit{Position} shape functions are, in general, prevailing in importance over the others. Comparing to other binary classifiers, both EBM and GAMI-Net placed in bewteen XGBoost and Logistic Regression models, reinforcing their position as a balance between predictive power and transparency.

A key limitation of using GAMs with interactions for extractive summarization is the absence of a mechanism to select good ``sets" of sentences, rather than just individually relevant sentences. Furthermore, the semantic information gap, arising from the challenge of incorporating it through non-dense features, could be a means of obtaining even more powerful models in the future.

\section{Conclusions}

In this work, we present the application of EBM and GAMI-Net to interpretable extractive summarization, as a simple but attractive alternative to traditional classification algorithms. Our results show that, despite more restrictive than full-complexity models in terms of formulation, GAMs with interactions were able to achieve  similar results to former black-box models.

Although the need for feature engineering can be seen as a disadvantage when comparing traditional approaches to neural models, with a concise set of features, both EBM and GAMI-Net models showed promising results for extractive summarization in textual datasets. The combination of intelligible features and the transparency of GAMs with interactions can be a tool to enlighten the view of the Extractive Summarization decisive process.

We present this paper as a preliminary effort concerning the topic of learning-based interpretable extractive summarization and believe that the perceptions presented into this work could help future research exploring the topic of intelligibility for ATS systems.

\section*{Acknowledgements}

The authors are grateful to FAPESP grants \#2013/07375-0, \#2014/12236-1, \#2019/07665-4, \#2019/18287-0, and \#2021/05516-1, and CNPq grant 308529/2021-9.

\bibliographystyle{splncs04}
\bibliography{references}

\begin{thebibliography}{10}
\providecommand{\url}[1]{\texttt{#1}}
\providecommand{\urlprefix}{URL }
\providecommand{\doi}[1]{https://doi.org/#1}

\bibitem{afsharizadeh2018query}
Afsharizadeh, M., Ebrahimpour-Komleh, H., Bagheri, A.: Query-oriented text
  summarization using sentence extraction technique. In: 2018 4th international
  conference on web research (ICWR). pp. 128--132. IEEE (2018)

\bibitem{arrieta2020explainable}
Arrieta, A.B., D{\'\i}az-Rodr{\'\i}guez, N., Del~Ser, J., Bennetot, A., Tabik,
  S., Barbado, A., Garc{\'\i}a, S., Gil-L{\'o}pez, S., Molina, D., Benjamins,
  R., et~al.: Explainable artificial intelligence (xai): Concepts, taxonomies,
  opportunities and challenges toward responsible ai. Information Fusion
  \textbf{58},  82--115 (2020)

\bibitem{bird2009natural}
Bird, S., Klein, E., Loper, E.: Natural language processing with Python:
  analyzing text with the natural language toolkit. " O'Reilly Media, Inc."
  (2009)

\bibitem{cohan2018discourse}
Cohan, A., Dernoncourt, F., Kim, D.S., Bui, T., Kim, S., Chang, W., Goharian,
  N.: A discourse-aware attention model for abstractive summarization of long
  documents. In: Proceedings of the 2018 Conference of the North American
  Chapter of the Association for Computational Linguistics: Human Language
  Technologies, Volume 2 (Short Papers). pp. 615--621 (2018)

\bibitem{danilevsky2020survey}
Danilevsky, M., Qian, K., Aharonov, R., Katsis, Y., Kawas, B., Sen, P.: A
  survey of the state of explainable ai for natural language processing. In:
  Proceedings of the 1st Conference of the Asia-Pacific Chapter of the
  Association for Computational Linguistics and the 10th International Joint
  Conference on Natural Language Processing. pp. 447--459 (2020)

\bibitem{dovsilovic2018explainable}
Do{\v{s}}ilovi{\'c}, F.K., Br{\v{c}}i{\'c}, M., Hlupi{\'c}, N.: Explainable
  artificial intelligence: A survey. In: 2018 41st International convention on
  information and communication technology, electronics and microelectronics
  (MIPRO). pp. 0210--0215. IEEE (2018)

\bibitem{el2020automatic}
El-Kassas, W.S., Salama, C.R., Rafea, A.A., Mohamed, H.K.: Automatic text
  summarization: A comprehensive survey. Expert Systems with Applications
  \textbf{165},  113679 (2020)

\bibitem{ferreira2013assessing}
Ferreira, R., de~Souza~Cabral, L., Lins, R.D., e~Silva, G.P., Freitas, F.,
  Cavalcanti, G.D., Lima, R., Simske, S.J., Favaro, L.: Assessing sentence
  scoring techniques for extractive text summarization. Expert systems with
  applications  \textbf{40}(14),  5755--5764 (2013)

\bibitem{ghodratnama2020extractive}
Ghodratnama, S., Beheshti, A., Zakershahrak, M., Sobhanmanesh, F.: Extractive
  document summarization based on dynamic feature space mapping. IEEE Access
  \textbf{8},  139084--139095 (2020)

\bibitem{hastie1987generalized}
Hastie, T., Tibshirani, R.: Generalized additive models: some applications.
  Journal of the American Statistical Association  \textbf{82}(398),  371--386
  (1987)

\bibitem{hermann2015teaching}
Hermann, K.M., Kocisky, T., Grefenstette, E., Espeholt, L., Kay, W., Suleyman,
  M., Blunsom, P.: Teaching machines to read and comprehend. Advances in neural
  information processing systems  \textbf{28} (2015)

\bibitem{kedzie2018content}
Kedzie, C., Mckeown, K., Daum{\'e}~III, H.: Content selection in deep learning
  models of summarization. In: Proceedings of the 2018 Conference on Empirical
  Methods in Natural Language Processing. pp. 1818--1828 (2018)

\bibitem{lin2004rouge}
Lin, C.Y.: Rouge: A package for automatic evaluation of summaries. In: Text
  summarization branches out. pp. 74--81 (2004)

\bibitem{liu2019fine}
Liu, Y.: Fine-tune bert for extractive summarization. arXiv preprint
  arXiv:1903.10318  (2019)

\bibitem{lou2012intelligible}
Lou, Y., Caruana, R., Gehrke, J.: Intelligible models for classification and
  regression. In: Proceedings of the 18th ACM SIGKDD international conference
  on Knowledge discovery and data mining. pp. 150--158 (2012)

\bibitem{lou2013accurate}
Lou, Y., Caruana, R., Gehrke, J., Hooker, G.: Accurate intelligible models with
  pairwise interactions. In: Proceedings of the 19th ACM SIGKDD international
  conference on Knowledge discovery and data mining. pp. 623--631 (2013)

\bibitem{luhn1958automatic}
Luhn, H.P.: The automatic creation of literature abstracts. IBM Journal of
  research and development  \textbf{2}(2),  159--165 (1958)

\bibitem{maybury1995generating}
Maybury, M.T.: Generating summaries from event data. Information Processing \&
  Management  \textbf{31}(5),  735--751 (1995)

\bibitem{montani2021spacy}
Montani, I., Honnibal, M., Honnibal, M., Landeghem, S.V., Boyd, A.: {spaCy:
  industrial-strength natural language processing in Python}  (2021),
  \url{https://doi.org/10.5281/zenodo.5648257}

\bibitem{moratanch2017survey}
Moratanch, N., Chitrakala, S.: A survey on extractive text summarization. In:
  2017 international conference on computer, communication and signal
  processing (ICCCSP). pp.~1--6. IEEE (2017)

\bibitem{mutlu2019multi}
Mutlu, B., Sezer, E.A., Akcayol, M.A.: Multi-document extractive text
  summarization: A comparative assessment on features. Knowledge-Based Systems
  \textbf{183},  104848 (2019)

\bibitem{nallapati2017summarunner}
Nallapati, R., Zhai, F., Zhou, B.: Summarunner: A recurrent neural network
  based sequence model for extractive summarization of documents. In:
  Proceedings of the Thirty-First AAAI Conference on Artificial Intelligence.
  p. 3075–3081. AAAI'17, AAAI Press (2017)

\bibitem{nenkova2011automatic}
Nenkova, A., McKeown, K.: Automatic summarization. Now Publishers Inc (2011)

\bibitem{nenkova2012survey}
Nenkova, A., McKeown, K.: A survey of text summarization techniques. In: Mining
  text data, pp. 43--76. Springer (2012)

\bibitem{nori2019interpretml}
Nori, H., Jenkins, S., Koch, P., Caruana, R.: Interpretml: A unified framework
  for machine learning interpretability. arXiv preprint arXiv:1909.09223
  (2019)

\bibitem{oliveira2016assessing}
Oliveira, H., Ferreira, R., Lima, R., Lins, R.D., Freitas, F., Riss, M.,
  Simske, S.J.: Assessing shallow sentence scoring techniques and combinations
  for single and multi-document summarization. Expert Systems with Applications
   \textbf{65},  68--86 (2016)

\bibitem{samek2020toward}
Samek, W., Montavon, G., Lapuschkin, S., Anders, C.J., M{\"u}ller, K.R.: Toward
  interpretable machine learning: Transparent deep neural networks and beyond.
  arXiv preprint arXiv:2003.07631  (2020)

\bibitem{samek2019towards}
Samek, W., M{\"u}ller, K.R.: Towards explainable artificial intelligence. In:
  Explainable AI: interpreting, explaining and visualizing deep learning, pp.
  5--22. Springer (2019)

\bibitem{sarkhel2020interpretable}
Sarkhel, R., Keymanesh, M., Nandi, A., Parthasarathy, S.: Interpretable
  multi-headed attention for abstractive summarization at controllable lengths.
  In: Proceedings of the 28th International Conference on Computational
  Linguistics. pp. 6871--6882 (2020)

\bibitem{see2017get}
See, A., Liu, P.J., Manning, C.D.: Get to the point: Summarization with
  pointer-generator networks. In: Proceedings of the 55th Annual Meeting of the
  Association for Computational Linguistics (Volume 1: Long Papers). pp.
  1073--1083 (2017)

\bibitem{wong2008extractive}
Wong, K.F., Wu, M., Li, W.: Extractive summarization using supervised and
  semi-supervised learning. In: Proceedings of the 22nd international
  conference on computational linguistics (Coling 2008). pp. 985--992 (2008)

\bibitem{wu2021r}
Wu, L., Li, J., Wang, Y., Meng, Q., Qin, T., Chen, W., Zhang, M., Liu, T.Y.,
  et~al.: R-drop: regularized dropout for neural networks. Advances in Neural
  Information Processing Systems  \textbf{34} (2021)

\bibitem{xiao2019extractive}
Xiao, W., Carenini, G.: Extractive summarization of long documents by combining
  global and local context. In: Proceedings of the 2019 Conference on Empirical
  Methods in Natural Language Processing and the 9th International Joint
  Conference on Natural Language Processing (EMNLP-IJCNLP). pp. 3011--3021
  (2019)

\bibitem{xiao2020systematically}
Xiao, W., Carenini, G.: Systematically exploring redundancy reduction in
  summarizing long documents. arXiv preprint arXiv:2012.00052  (2020)

\bibitem{yang2021gami}
Yang, Z., Zhang, A., Sudjianto, A.: Gami-net: An explainable neural network
  based on generalized additive models with structured interactions. Pattern
  Recognition  \textbf{120},  108192 (2021)

\bibitem{zhong2020extractive}
Zhong, M., Liu, P., Chen, Y., Wang, D., Qiu, X., Huang, X.J.: Extractive
  summarization as text matching. In: Proceedings of the 58th Annual Meeting of
  the Association for Computational Linguistics. pp. 6197--6208 (2020)

\end{thebibliography}

\end{document}